\documentclass[letterpaper, 10 pt, conference]{ieeeconf}  % Comment this line out if you need a4paper

\IEEEoverridecommandlockouts                              % This command is only needed if 
\usepackage[nolist,nohyperlinks]{acronym}

\usepackage{todonotes}
\usepackage{graphicx}
\usepackage{caption}
\usepackage{subfig}

\usepackage{amsmath}
\usepackage{amssymb}
\usepackage{tabularx}
\usepackage{paralist}
\usepackage{algorithm}
\usepackage[noend]{algpseudocode}

\newcommand{\norm}[1]{\vert\vert #1 \vert\vert }
\makeatletter
\def\BState{\State\hskip-\ALG@thistlm}
\makeatother

\title{\LARGE \bf Collaborative Object Transportation Using MAVs via Passive Force Control}

\author{$\text{Andrea Tagliabue}^{1}$, $\text{Mina Kamel}^{1}$, $\text{Sebastian Verling}^{1}$, $\text{Roland Siegwart}^{1}$, $\text{Juan Nieto}^{1}$\thanks{${}^1$ Autonomous Systems Lab., Department of Mechanical and Process Engineering, ETH Zurich, Zurich, Switzerland. }% <-this % stops a space
}

\begin{document}
\maketitle
\thispagestyle{empty}
\pagestyle{empty}

%%%%%%%%%%%%%%%%%%%%%%%%%%%%%%%%%%%%%%%%%%%%%%%%%%%%%%%%%%%%%%%%%%%%%%%%%%%%%%%%
\begin{abstract}
This paper shows a strategy based on passive force control for collaborative object transportation using \acp{MAV}, focusing on the transportation of a bulky object by two hexacopters. The goal is to develop a robust approach which does not rely on:
\begin{inparaenum}[(a)]
 \item communication links between the \acp{MAV},
 \item the knowledge of the payload shape and
\item the position of grasping point.
\end{inparaenum}The proposed approach is based on the master-slave paradigm, in which the slave  agent guarantees compliance to the external force applied by the master to the payload via an admittance controller. The external force acting on the slave is estimated using a non-linear estimator based on the \ac{UKF} from the information  provided by a visual inertial navigation system. Experimental results demonstrate the performance of the force estimator and show the collaborative transportation of a $1.2$ m long object.  
\end{abstract}

%%%%%%%%%%%%%%%%%%%%%%%%%%%%%%%%%%%%%%%%%%%%%%%%%%%%%%%%%%%%%%%%%%%%%%%%%%%%%%%%
% Introduction
%%%%%%%%%%%%%%%%%%%%%%%%%%%%%%%%%%%%%%%%%%%%%%%%%%%%%%%%%%%%%%%%%%%%%%%%%%%%%%%%

\section{INTRODUCTION}
\acp{MAV} have recently stimulated the fantasy of researchers and entrepreneurs \cite{ambulance_drone_2016}\cite{zipline} as a new tool for good delivery. Quadcopters and hexacopters have shown to be especially suited for this task, thanks to their ability to navigate in cluttered environments \cite{Nieuwenhuisen2015} and being able to deliver a payload with extreme accuracy \cite{Architecture2014}. However, their inherent limited size constrain the range of applications to the transportation of small and light-weight objects.
\par
Collaborative strategies can significantly enhance \ac{MAV} transportation capabilities and can provide a cost effective solution with respect to the deployment of a single, more capable \ac{MAV} \cite{Maza2010}. A common approach for collaborative transportation is based on centralized solutions, where a coordinator computes a control action for each of the agents and it shares the command with them \cite{Michael2010}. A second possible approach is based on distributed control algorithm. In this case, the agents only share a common goal and the individual control laws are obtained with respect to the grasping point on the payload \cite{Mellinger2015}.
\par
Both of the described collaborative approaches, however, present some limitations. Centralized approaches rely on the ability of the coordinator to effectively communicate with the agents and this can often be an issue due to the limited robustness of wireless communication networks. Distributed approach, instead, relies on the knowledge of the relative position between each agent and the payload. %This information, anyway, can not be trivially retrieved if the grasping point is not predetermined and the shape and size of the object are not known a-priori.
\par 
In this paper we overcame these issues by showing a bio-inspired \cite{McCreery2014} collaborative approach based on the master-slave(s) paradigm. We chose one of the available \acp{MAV} to be the leader - the master - while the remaining agents act as slaves. The task of the master is simply to lift the payload and pull it in the desired direction. The slaves - which are also attached to the payload - actively guarantee compliance to the master's actions by sensing the force that master applies on the payload and changing their position accordingly. Slave's active compliance is guaranteed through an admittance controller, while external forces are estimated via position, velocity, attitude and angular velocity information using an estimator based on the \ac{UKF}. To obtain the pose of the \ac{MAV}, as well as its velocity and angular velocity, we use a visual-inertial navigation system mounted on the slave agent. 
\par In our setup we assume the following: (a) the payload is big enough to guarantee that at least two \acp{MAV} can autonomously grasp it, (b) all the agents are already attached to the payload and (c) the attitude of the vehicle is decoupled from the rotational dynamic of the payload - this is achieved, for example, by connecting vehicle and payload via a rope or a spherical joint \cite{Nguyen2015}.
\begin{figure}[t]
\centering
\includegraphics[width=0.47\textwidth]{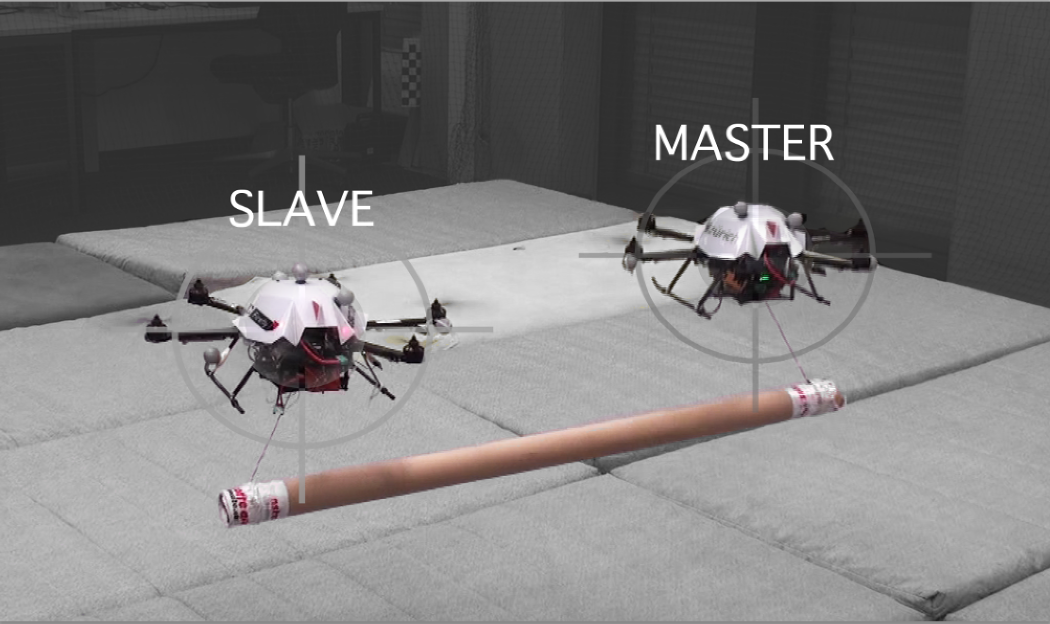}
  	\caption{Experimental setup for collaborative object transportation. The master vehicle is controlled remotely, while the slave provides compliance to the movement of the master using an admittance controller.}
\label{pic:exp:masterslave}
\end{figure}
\subsection{Contributions}
%Try to make this statement less qualitative and more quantitative (confutable)
The main contribution of this work is the novel approach for collaborative object transportation using \acp{MAV}. The approach that we propose has the advantage of not relying on:  
\begin{inparaenum}[(a)]
 \item communication links between the \acp{MAV},
 \item the knowledge of the payload shape, 
\item the position of grasping point,
\end{inparaenum} and is potentially scalable to multiple agents.
To effectively demonstrate our approach, we describe the implementation of a force estimator based on the \ac{UKF} which handles quaternions, is computationally light, and fast. It is able to detect a collision with a wall in less than 20 ms and provides an accurate enough force estimate for human-vehicle interaction without relying on an external motion capture system, but using a VI (Visual Inertial) navigation system.

\subsection{Structure of this work}
We start by presenting existing approaches and technologies for cooperative transportation via \ac{MAV}  in  Section \ref{sec:relwork}. We then proceed by providing an overview of the control strategy and architecture in Section \ref{sec:sysdes} and we detail the implementation of the force estimator and admittance controller in Sections \ref{sec:ukf} and \ref{sec:admc}, respectively. We conclude by showing experimental results in Section \ref{sec:exp} and conclusion in Section \ref{sec:futwrk}.

%%%%%%%%%%%%%%%%%%%%%%%%%%%%%%%%%%%%%%%%%%%%%%%%%%%%%%%%%%%%%%%%%%%%%%%%%%%%%%%%
% Related work
%%%%%%%%%%%%%%%%%%%%%%%%%%%%%%%%%%%%%%%%%%%%%%%%%%%%%%%%%%%%%%%%%%%%%%%%%%%%%%%%

\section{RELATED WORK}
\label{sec:relwork}
Literature for collaborative transportation via aerial vehicle is rich of examples. Earliest publications are related to collaborative transportation of a single payload using two helicopters \cite{Mittal1991} \cite{Reynolds1992}, but do not provide experimental results. Collaborative transportation using UAVs are extensively discussed by \cite{Maza2010} and \cite{Parra-Vega2013}.  \cite{Michael2010} makes use of a centralized cooperative strategy which relies on communication between \acp{MAV} while an example of distributed strategy based on information about the payload shape and grasping points is provided by \cite{Mellinger2015}.
\par
In terms of passive force control and admittance control, a comprehensive description is provided in \cite{Handbook2008}. 
\cite{Augugliaro2013} describes an example of admittance control for quadcopters and human-machine interaction, but heavily relies on motion capture systems for the force estimation.
\par An external forces and torque estimator for multirotor vehicles based on the UKF and similar to the one we propose is adressed on \cite{Mckinnon2016}, which relies on the work of \cite{Crassidis2003} for attitude quaternion estimation. 
%In addition: we use quaternion averaging algorithm, we include aerodynamics effect in force model, we assume zero torque around x and y axis.
%\cite{Maza2010}, \cite{Parra-Vega2013}
%Aherial Vehicles \cite{Mittal1991}
%Cooperative solution \cite{Michael2010}
%Distributed solution \cite{Mellinger2015}
%admittance control \cite{Augugliaro2013} \cite{Handbook2008}
%unscented kalman filter \cite{Mckinnon2016}, \cite{Crassidis2003}

%http://spectrum.ieee.org/automaton/robotics/drones/defibrillator-drone-another-good-drone-idea

%%%%%%%%%%%%%%%%%%%%%%%%%%%%%%%%%%%%%%%%%%%%%%%%%%%%%%%%%%%%%%%%%%%%%%%%%%%%%%%%
% Method
%%%%%%%%%%%%%%%%%%%%%%%%%%%%%%%%%%%%%%%%%%%%%%%%%%%%%%%%%%%%%%%%%%%%%%%%%%%%%%%%

\section{SYSTEM DESIGN}
\label{sec:sysdes}
\begin{figure*}
	\centering
	\includegraphics[width=\textwidth]{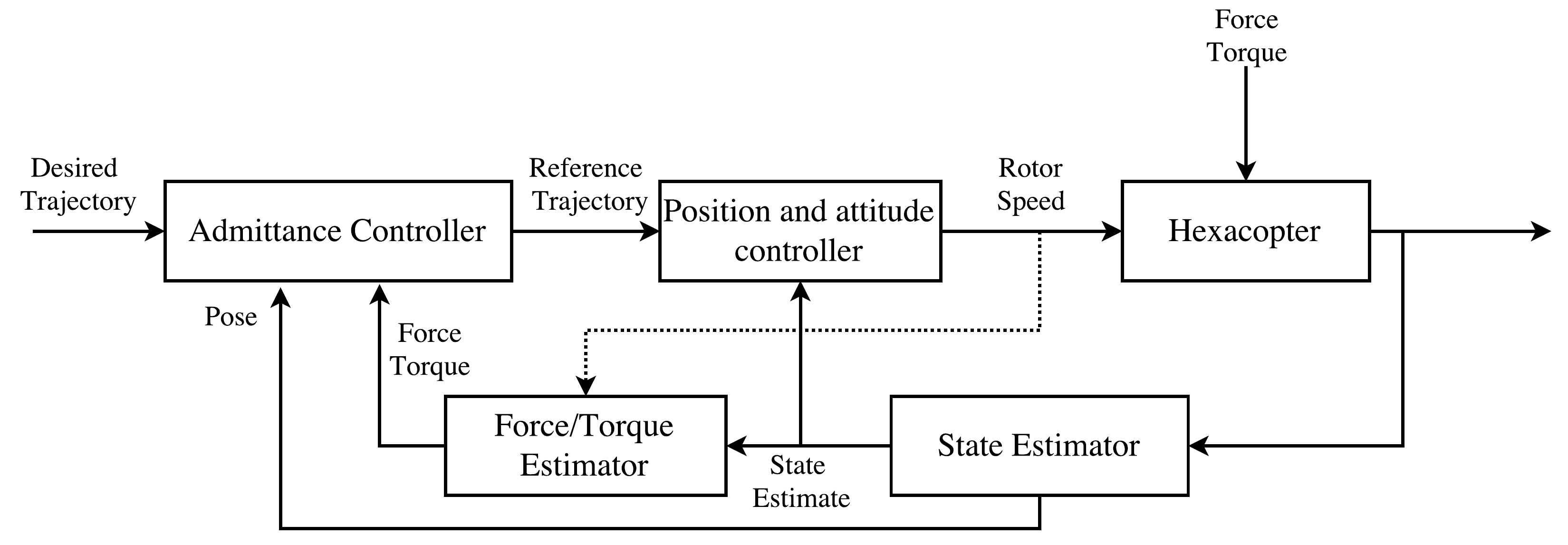}
	\caption{Active interaction control scheme for the slave agent. The user-provided desired trajectory is modified by the admittance controller according to the external force and torque estimated by the \ac{UKF} filter based estimator. The state estimator can be either based on a visual-inertial navigation system or an external motion capture system.}
	\label{pic:architecture:slave}
\end{figure*}
\subsection{The Master-Slave Paradigm}
The master-slave paradigm is a common communication approach for cooperative tasks and is used in many different fields, from industries, informatics to the railway sector. In our context, we defined it as follows. We divide the agents (the \acp{MAV}) which take part in the collaborative transportation maneuver in two categories, (a) master and (b) slave. The master, which is unique, follows a reference trajectory given by an external operator or a path planning algorithm while grasping, lifting and pulling the payload. It executes its standard on-board reference tracking and state estimation algorithms and it behaves as if it was carrying the payload alone. The slave agent - which can be one or more - instead grasps the payload, detects the magnitude and direction of the force that the master is applying on it and generates a trajectory compliant with the force applied by the master. \par
Master and slave do not share information via a conventional communication link, but rather via the force applied to the payload, which is sensed by the slave through a force estimator.

\subsection{Slave Control Architecture}
From now on we will focus our attention on the control architecture of the slave agent, since the master runs a standard pose tracking  feedback loop. Slave's control architecture is composed by four main building blocks.
\begin{itemize}
\item \textbf{State estimator}: 
estimates the slave's
\begin{inparaenum}[(a)]
\item position,
\item velocity,
\item attitude, and 
\item angular velocity
\end{inparaenum}
w.r.t the \textit{I} frame. It can either be an external motion capture system such as \cite{vicon} or a VI navigation system. The attitude is expressed as a quaternion.
\item \textbf{Force and torque estimator}: estimates the external force acting on the slave expressed in \textit{I} frame and external torque around $z_B$ axis in \textit{B} frame. Additionally to the full output of the state estimator - which is used in the update step - the force and torque estimator needs the measurement of the speed of the rotors to propagate the model dynamics in the state prediction step.
\item \textbf{Admittance controller}: provides a reference pose (position and attitude) for the \ac{MAV} position and attitude controller, given the estimate of the external force and torque and a desired trajectory. According to the choice of its  parameters, it can better track the desired trajectory or give full compliance to the estimated external disturbances. For example, during the cooperative transportation, it complies fully to the force components on $x_I$ and $y_I$ axis while tracking an altitude reference on $z_i$. More details can be found in Section \ref{sec:admc}. 
\item \textbf{Position and attitude controller}: MPC based controller, that tracks the trajectory generated by the admittance controller by providing a rotor speed command.
\end{itemize}

\subsection{Coordinate System}
\begin{figure}
\centering
\includegraphics[width=0.3\textwidth]{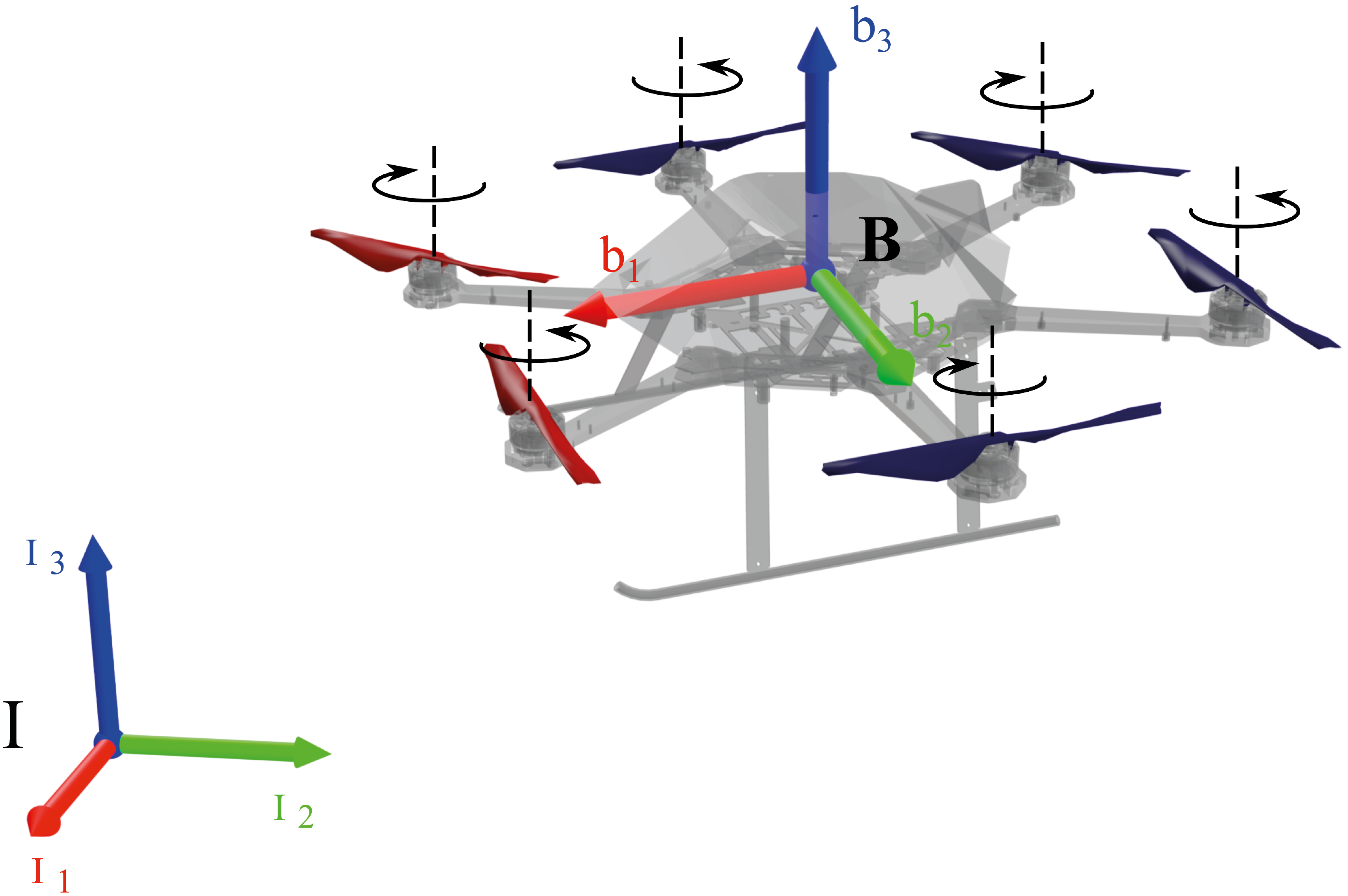}
  	\caption{Reference frames defined for this work. \textit{I} represents the inertial reference frame, while \textit{B} the \ac{MAV} reference frame.}
\label{pic:sys_des:ref_frame}
\end{figure}
The relevant coordinate frames for this work are two: an inertial reference frame attached to the ground $I$ and a non-inertial reference frame attached to the Center of Gravity (CoG)of the vehicle $B$. They are represented in Figure \ref{pic:sys_des:ref_frame}.

\subsection{Hardware and Software}
\paragraph{Hardware}
The \ac{MAV} used for our experiments is the \textbf{AscTec Firefly hexacopter} equipped with an on board computer based on a \textbf{quad-core 2.1 GHz Intel i7} processor and \textbf{8 GB of RAM}. The hexacopter is additionally equipped with a Visual-Inertial navigation system, developed by the Autonomous Systems Lab at ETHZ and Skybotix AG \cite{J.Nikolic2014}. This sensor measurement, fused with vehicle's IMU provide a 100 Hz estimate of:
\begin{inparaenum}[(a)]
\item position, 
\item velocity, 
\item attitude,
\item angular velocity
\end{inparaenum}
of the \ac{MAV}.
\paragraph{Software}
We implemented the force estimator and the admittance controller algorithm using \textbf{C++ and ROS}. All the algorithms run on board the hexacopter to avoid issues related to wireless communication. Both the force estimator and the admittance controller run at \textbf{100 Hz} using about 10\% of the available computational power together. The position and attitude controller is based on the model predictive controller developed in \cite{kamelmpc2016,2016arXiv161109240K}.

%%%%%%%%%%%%%%%%%%%%%%%%%%%%%%%%%%%%%%%%%%%%%%%%%%%%%%%%%%%%%%%%%%%%%%%%%%%%%%%%
% Implementation
%%%%%%%%%%%%%%%%%%%%%%%%%%%%%%%%%%%%%%%%%%%%%%%%%%%%%%%%%%%%%%%%%%%%%%%%%%%%%%%%

\section{EXTERNAL FORCE AND TORQUE ESTIMATOR}
\label{sec:ukf}
In this section, we present the \ac{UKF} that was implemented to estimate the external force and torque acting on the \ac{MAV}.  First, we derive a nonlinear, discrete time model of the rotational and translational dynamics of a hexacopter, to be used in the derivation of the process model of the filter. Second, we derive the measurement model to be used in the update step. Third, we describe the prediction and update step for the whole state of the filter with the exception of the attitude quaternion. In the end, we describe the singularity free prediction and update steps which take into account the attitude quaternion.
\subsection{Hexacopter Model with External Force and Torque}
\paragraph{Model Assumptions}
In order to simplify the model derivation, we assume that: 
\begin{itemize}
	\item The \ac{MAV} structure is rigid, symmetric on the ${x_{B}y_{B}}$ and $y_{B}z_{B}$ planes, and the CoG (Center of Gravity) and body frame origin \textit{B} coincide;
	\item Aerodynamic interaction with the ground or other surfaces can be neglected and we only consider aerodynamic effects dominant at low speed;
	\item No external torque acts around $x_B$ and $y_B$ axis.
\end{itemize}

\par
In addition, we define the rotation matrix from frame $B$ to frame $I$ as 
$\boldsymbol{R(q)}_{B}^{I}$, where $\boldsymbol{q}$ is the normalized quaternion representing the attitude of the vehicle. Finally, we simplify the equations by introducing $U_{1}$, $U_{2}$ and $U_{3}$ as the total torque produced by the propellers around, respectively, $x_b$, $y_b$ and $z_b$, and $U_{4}$ as the total thrust produced by the propellers expressed in $B$ frame. We compute $U_{i}, i = 1\dots4$ from the rotor speed $n_{i}$, $i=1\dots6$, using the allocation matrix, as defined in \cite{Achtelik2013} for the AscTec Firefly hexacopter \cite{asctec}:

 \begin{equation}
 \begin{bmatrix}
 U_{1} \\ U_{2} \\ U_{3} \\ U_{4} \\
 \end{bmatrix}
 =
 \mathbf{K}
 \begin{bmatrix}
  s & 1 & s & -s & -1 & s \\
-c & 0 & c & c & 0 & -c \\
-1 & 1 & -1 & 1 & -1 & 1 \\
 1 & 1 & 1 & 1 & 1 & 1
 \end{bmatrix}
 \begin{bmatrix}
 {n_{1}}^2 \\
 \vdots \\
 {n_{6}}^2\\
 \end{bmatrix}
 \end{equation}
 
 \begin{equation}
  \mathbf{K} = 
  \text{diag}
   \big(
    \begin{bmatrix}
    l k_n& l k_n& k_n k_m& k_n
    \end{bmatrix}
   )
 \end{equation} 
where $l$ is the boom length, $k_n$ and $k_m$ are rotor constants and $c = \text{cos}(30\deg)$ and $s = \text{sin}(30\deg)$.
 
\paragraph{Translational Dynamics}
We derive the linear acceleration with respect to $I$ reference frame by taking into account the following forces: (a) gravity, (b) thrust, (c) aerodynamic effects, (d) external forces due to interaction with the vehicle. We obtain:
\begin{equation}
	\begin{bmatrix}
		\ddot{x} \\
		\ddot{y} \\
		\ddot{z}
	\end{bmatrix} 
	= \frac{1}{m}\boldsymbol{R(q)}_{B}^{I} \Bigg(
	\begin{bmatrix}
	0\\
	0\\
	U_{4}\\
	\end{bmatrix}
	+\boldsymbol{ F}^{aero}
	\Bigg) 
	-
	\begin{bmatrix}
	0\\0\\g
	\end{bmatrix}
		+ 
	\frac{1}{m} \boldsymbol{F}^{ext}
	\label{transldyn}
\end{equation}
where $\boldsymbol{F}^{ext}$ are the external forces that act on the vehicle expressed in $I$ reference frame, $m$ is the mass of the vehicle, and $\boldsymbol{F}^{aero}$ is defined as: 
\begin{equation}
\boldsymbol{F}^{aero} = k_{drag} \sum_{i=1}^{6}|n_{i}|
\begin{bmatrix}
\dot{x} \\
\dot{y} \\
0
\end{bmatrix}
\end{equation}

\paragraph{Rotational Dynamics}
We derive the angular acceleration $\dot{\boldsymbol{\omega}}$ in $B$ frame by taking into account the following torques: (a) total torque produced by propellers, (b) external torque around z axis, (c) inertial effects, (d) propeller's gyroscopic effects. We obtain:
\begin{equation}
\boldsymbol{\dot{\omega}} = \mathbf{J}^{-1}
\Bigg(
\begin{bmatrix} 
U_{1} \\
U_{2} \\
U_{3} \\
\end{bmatrix}
+
\begin{bmatrix}
0\\
0\\
\tau^{ext}
\end{bmatrix}
- \boldsymbol{\omega }\times \mathbf{J}\boldsymbol{\omega} 
- \boldsymbol{\tau^{rotor}}
\Bigg) \label{rotdyn}
\end{equation}
where $\tau_{ext}$ is the external torque around the $z_B$ axis.
$ \mathbf{J}$ is the inertia tensor of the \ac{MAV} with respect to  $B$  frame, defined as
\begin{equation}
\mathbf{J} = 
\begin{bmatrix}
J_{xx} & 0 & 0 \\
0 & J_{yy} & 0 \\
0 & 0 & J_{zz} 
\end{bmatrix}
\end{equation}
and $\tau_{rotor}$ is the torque caused by propeller's gyroscopic effects defined according to \cite{Bouabdallah2007} as
\begin{equation}
\boldsymbol{\tau_{rotor}} = J_{r}\Omega_{r}
\begin{bmatrix}
\omega_{y}\\
\omega_{x}\\
0
\end{bmatrix}
\textrm{with } 
\Omega_{r}  = \sum_{i = 1}^{6} n_{i} - 6 \omega_z
\end{equation}
where $J_r$ is the propeller's moment of inertia around its axis of rotation.

%%% DISCRETIZATION %%% 
\paragraph{Discretized Model}
Given the sampling time $T_s = 0.01$s, we discretize the analytic models (\ref{transldyn}) and (\ref{rotdyn}) using the forward Euler method, while the normalized attitude quaternion $\boldsymbol{q}_k$ is integrated using the approach proposed in \cite{Crassidis2003}:
\begin{equation}
\label{d_att}
\boldsymbol{q}_{k+1} = \boldsymbol{\Omega}(\boldsymbol{\omega}_k) \boldsymbol{q}_k
\end{equation}
with 
\begin{equation}
\boldsymbol{\Omega}(\boldsymbol{\omega}_k) = 
\begin{bmatrix}
\cos(\frac{1}{2}\norm{\boldsymbol{\omega}_k} T_s) \mathbf{I}_{3 \times 3}  & \mathbf{\Psi}_k \\
-\mathbf{\Psi}_k^T &  \cos(\frac{1}{2}\norm{\boldsymbol{\omega}_k} T_s)  
\end{bmatrix} 
\end{equation}
and 
\begin{equation}
\mathbf{\Psi}_k = \sin{(\frac{1}{2} \norm{\boldsymbol{\omega_k}}T_s)}\boldsymbol{\omega_k}/\norm{\boldsymbol{\omega_k}}
\end{equation}
The discretized dynamic equations can then be written in compact form as:
\begin{equation}
\label{discr:dynold}
\boldsymbol{{s}}_{k+1}= f(\boldsymbol{{s}}_k, \boldsymbol{F}_k^{ext}, 	\tau_k^{ext}, \boldsymbol{n}_{k})
\end{equation}
with 
\begin{equation}
\label{discr:state}
\boldsymbol{{s}}_k =
\begin{bmatrix}
\boldsymbol{p}_k&\boldsymbol{v}_k & \boldsymbol{q}_k & \boldsymbol{\omega}_k
\end{bmatrix}
\end{equation}
As the last step we introduce the notation:
\begin{equation}\label{discr:dyn}
\boldsymbol{s}_{k+1} = f_k(\boldsymbol{s}_k, \boldsymbol{F}_k^{ext}, 	\tau_k^{ext})
\end{equation}
where the dependency of $f(\cdot)$ from $\boldsymbol{n}_{k}$ is taken into account by making $f(\cdot)$ time-variant.

\subsection{Process and Measurement models}
\paragraph{Process Model}
First, we augment the state vector of the hexacopter model (\ref{discr:state}) and the system dynamic equation (\ref{discr:dyn}) so that they take into account  the external force and torque.
We obtain:
\begin{equation}
{\boldsymbol{s}}_{k+1} = {f}_k({\boldsymbol{s}}_k) + \bar{\boldsymbol{w}}_k
\end{equation}
\begin{equation} \label{ukf:pred:sysdyn}
\boldsymbol{s}_k = 
\begin{bmatrix}
\boldsymbol{p}_k&\boldsymbol{v}_k&\boldsymbol{q}_k&\boldsymbol{\omega}_k&\boldsymbol{F}_k^{ext}&\tau_k^{ext} 
\end{bmatrix}
\end{equation}
We assume that the change of external force and torque is purely driven by zero mean additive process noise, whose covariance is a tuning parameter of the filter. The external force and torque are updated according to 
\begin{equation}
\begin{aligned}
&\boldsymbol{F}_{k+1}^{ext} = \boldsymbol{F}_{k}^{ext} \\ 
& \tau_{k+1}^{ext} = \tau_{k}^{ext} \\
\end{aligned}  
\end{equation}
or by assuming an exponential force decaying model  
\begin{equation}
\begin{aligned}
&\boldsymbol{F}_{k+1}^{ext} = \boldsymbol{F}_{k}^{ext}(1- T_s/\bar{\tau}) \\ 
& \tau_{k+1}^{ext} = {\tau}_k^{ext}(1- T_s/\bar{\tau}) \\
\end{aligned}  
\end{equation}
where $\bar{\tau}$ is a constant set by the user and defines the convergence speed of the estimates. 
In order to define the state covariance and the process noise covariance, we introduce a new representation $\bar{\boldsymbol{s}}_{k}$ of the state ${\boldsymbol{s}}_{k}$, where we substitute the attitude quaternion $\boldsymbol{q}_k$ with a $3 \times 1$ attitude error vector $\boldsymbol{e}_k$. More details about this substitution are explained in \ref{ukf:attitude}.
\begin{equation}
\bar{\boldsymbol{s}}_k = 
\begin{bmatrix}
\boldsymbol{p}_k&\boldsymbol{v}_k&\boldsymbol{e}_k& \boldsymbol{\omega}_k& \boldsymbol{F}_k^{ext}& \tau_k^{ext} \\
\end{bmatrix}
\end{equation}
 We can now introduce $\boldsymbol{P}_k$, the $16 \times 16$ covariance matrix associated to the state $\bar{\boldsymbol{s}}_k$, and $\boldsymbol{Q}$, the $16 \times 16$ time invariant  process noise diagonal covariance matrix.

%The time invariant $16 \times 16 $ process noise diagonal covariance matrix is
%\begin{equation}
%\boldsymbol{Q} =  \text{diag}(
%\begin{bmatrix}
%\boldsymbol{1}_3^T \sigma_{p,p}^2& 
%\boldsymbol{1}_3^T \sigma_{p,v}^2&
%\boldsymbol{1}_3^T \sigma_{p,e}^2& 
%\boldsymbol{1}_3^T  \sigma_{p,\omega} ^2& 
%{\sigma_{p,F}^{xy}}^2& {\sigma_{p,F}^{xy}}^2& %{\sigma_{p,F}^{z}}^2& \sigma_{p,\tau}^2
%\end{bmatrix}
%)
%\end{equation}
%where $\sigma_{p,e}$ is the process uncertainty associated to the 3 states attitude error vector (see \ref{ukf:attitude}).

\paragraph{Measurement Model}
We define the measurement vector $\boldsymbol{z}_k$, the  associated $12 \times 12$ measurement noise covariance matrix $\boldsymbol{R}$ and the linear measurement function 
$\boldsymbol{z}_k  = \boldsymbol{H}{\bar{\boldsymbol{s}}}_k + \boldsymbol{v}_k $, under the assumption that the measurements are subject to zero-mean additive noise.
\begin{equation}
\boldsymbol{z}_k = 
\begin{bmatrix}
\label{ukf:meas_vect}
\boldsymbol{p}_k^m& \boldsymbol{v}_k^m& \boldsymbol{e}_k^m& 	\boldsymbol{\omega}_k^m
\end{bmatrix}
\end{equation}

\begin{equation} \label{ukf:meas_funct}
\begin{aligned}
\boldsymbol{z}_k & = \boldsymbol{H}{\bar{\boldsymbol{s}}}_k + \boldsymbol{v}_k \\
\boldsymbol{H} =  &
\begin{bmatrix}	
\mathbf{I}_{12\times12} & \mathbf{0}_{12 \times 4}
\end{bmatrix}
\end{aligned}
\end{equation}
\begin{equation} \label{ukf:meas_cov}
\boldsymbol{R} = 
\text{diag}(
\begin{bmatrix}
\boldsymbol{1}_3^T \sigma_{m,p}^2& 
\boldsymbol{1}_3^T \sigma_{m,v}^2&
\boldsymbol{1}_3^T \sigma_{m,e}^2& 
\mathbf{1}_3^T  \sigma_{m,\omega} ^2
\end{bmatrix}
)
\end{equation}

As a remark, we observe that the measured attitude quaternion is converted in a three states parametrization $\boldsymbol{e}_k$ for the measurement vector. In addition, the measurement update step of the filter is performed using the state vector $\bar{\boldsymbol{s}}_k$. (\ref{ukf_simpl_updt}).

\subsection{Position, Velocity, Angular Velocity, External Force and Torque Estimation}

\textbf{Notation Declaration:}
From now on, we will use the following notation:
\begin{itemize}
	\item $\hat{\boldsymbol{s}}_{k-1}^{+}$ denotes the estimated state before the prediction step
	\item $\hat{\boldsymbol{s}}_{k}^{-}$ denotes the estimated state after the prediction step and before the update step
	\item $\hat{{\boldsymbol{s}}}_{k}^{+}$ denotes the estimated state after the update step
\end{itemize}
The same applies for the state covariance matrix  $\boldsymbol{P}$.

\paragraph{Prediction Step}
The predicted state for all the states with the exception of the attitude is computed using the standard \ac{UKF} prediction step, as explained in \cite{Simon2006} and \cite{Julier1996}. The matrix square root is computed using the Cholesy decomposition \cite{Simon2006}.

\paragraph{Kalman Filter Based Update Step}
\label{ukf_simpl_updt}
Because the measurement model corresponds to a linear function, we can use the standard KF update step for all the states in the vector state $\bar{\boldsymbol{s}}_k$. 
\par First, compute the Kalman gain matrix as 
\begin{equation}
\boldsymbol{K}_k = \boldsymbol{P}_k^- \boldsymbol{H}^T (\boldsymbol{H} \boldsymbol{P}_k^- \boldsymbol{H}^T + \boldsymbol{R})^{-1}
\end{equation}

Then, the updated state covariance  is obtained from 
\begin{equation}
\boldsymbol{P}_k^+ = (\mathbf{I}_{16\times16} - \boldsymbol{K}_k \boldsymbol{H}) \boldsymbol{P}_k^- (\mathbf{I}_{16\times16} - \boldsymbol{K}_k \boldsymbol{H})^T + \boldsymbol{K}_k \boldsymbol{R} \boldsymbol{K}_k^T
\end{equation}

Finally, the updated state is obtained from
\begin{equation}
\bar{\boldsymbol{s}}_k^+= \bar{\boldsymbol{s}}_k^- + \boldsymbol{K}_k(\boldsymbol{z}_k - \boldsymbol{H}\bar{\boldsymbol{s}}_k^- ) 
\end{equation}

\subsection{Quaternion Based Attitude Estimation}
\label{ukf:attitude}
Quaternions do not naturally fit in the Unscented Transformation (UT) since they are defined on a non linear manifold, while the UT is performed on a vector space. Interpreting a quaternion as a member of $\mathbb{R}^4$ produces a singular $4 \times 4$ covariance matrix and does not  guarantee the unitary norm constraint, while considering quaternions as members of $\mathbb{R}^3$ - by exploiting the unitary norm constraint - always produces a singularity. For these reasons we estimate the \ac{MAV} attitude by following the approach called USQUE - Unscented Quaternion Estimation - proposed by \cite{Crassidis2003}. In this approach, the attitude estimate is based on the estimation of the error attitude quaternion - which is always assumed to be smaller than 180 degrees -  through a three states parametrization obtained using the Modified Rodriguez Parameters (MRP). 

\textbf{Modified Rodriguez Parameters}
Let $\boldsymbol{q} = [\boldsymbol{q}_v^T, {q}_s]$ be a quaternion, where $\mathbf{\boldsymbol{q}_v}$ is the vector part and $q_s$ is the scalar part. A corresponding (but singular) attitude representation $\mathbf{p}$, with $\mathbf{p} \in \mathbb{R}^3$, can be obtained as
\begin{equation} \label{mrp:fromquat}
\mathbf{p} = f \frac{\mathbf{\boldsymbol{q}_v}}{a+q_s} 
\end{equation}
where $a$ is a parameter from 0 to 1 and f is a scale factor. We will choose $f = 2(a+1)$ and $a$ to be a tunable parameter of the filter. The inverse representation is obtained as 
\begin{equation} 
\label{mrp:toquat}
\begin{aligned}
q_s =& \frac{-a\ \norm{\mathbf{p}}^2 + f \sqrt{f^2+(1-a^2) \norm{\mathbf{p}}^2}}{f^2+\norm{\mathbf{p}}^2} \\
\mathbf{\boldsymbol{q}_v} =& f^{-1}(a+q_s) \mathbf{p}
\end{aligned}
\end{equation}

The implemented algorithm for attitude estimation is discussed in the following steps. Its prediction step follows the USQUE approach explained in \cite{Crassidis2003}. The update step - where we fuse the attitude measurement at time $k$ into the attitude error vector - assumes that the measurement for the filter corresponds to the difference between the measured attitude quaternion and the estimate of the attitude at time step $k-1$. Because the initialization and prediction step are extensively covered in \cite{Crassidis2003} and \cite{Markley2007}, we only report the implemented measurement update step.

	\paragraph{Update Step}
	\begin{itemize}
		\item Given an attitude measurement $\boldsymbol{q}_k^m$, compute the rotation error between $\hat{\boldsymbol{q}}_{k-1}^+$ and $\boldsymbol{q}_k^m$ (\ref{ukf:att:meas}).
		\begin{equation} \label{ukf:att:meas}
		\delta \boldsymbol{q}_k^m = \hat{\boldsymbol{q}}_{k-1}^+ \otimes {(\boldsymbol{q}_k^m)}^{-1}
		\end{equation}
		\item Transform the measured rotation error quaternion $\delta \boldsymbol{q}_k^m$ in the vector  $\boldsymbol{e}_k^m$ using MRP (\ref{mrp:fromquat}). 
		\item Store $\boldsymbol{e}_k^m$ in $\bar{\hat{\boldsymbol{s}}}_k^-$ and compute the updated attitude error mean and covariance as in \ref{ukf_simpl_updt}. 
		\item Retrieve $\hat{\boldsymbol{e}}_k^+$ from $\bar{\hat{\boldsymbol{s}}}_k^+$ and compute the updated error attitude quaternion $\delta \boldsymbol{q}_k^+$ from $\hat{\boldsymbol{e}}_k^+$ using MRP (\ref{mrp:toquat}).
		\item Rotate the predicted attitude quaternion $\hat{\boldsymbol{q}}_k^-$ of $\delta \boldsymbol{q}_k^+$ to obtain the current quaternion estimate of the attitude
		\begin{equation}
		\hat{\boldsymbol{q}}_k^+ = \hat{\boldsymbol{q}}_k^- \otimes \delta \boldsymbol{q}_k^+
		\end{equation}
	\end{itemize}

\section{ADMITTANCE CONTROLLER}
\label{sec:admc}
\subsection{An Intuitive Explanation}
\label{sec:admc:intuitive_explanation}
\begin{figure}[t]
	\centering
	\includegraphics[width=0.5\textwidth]{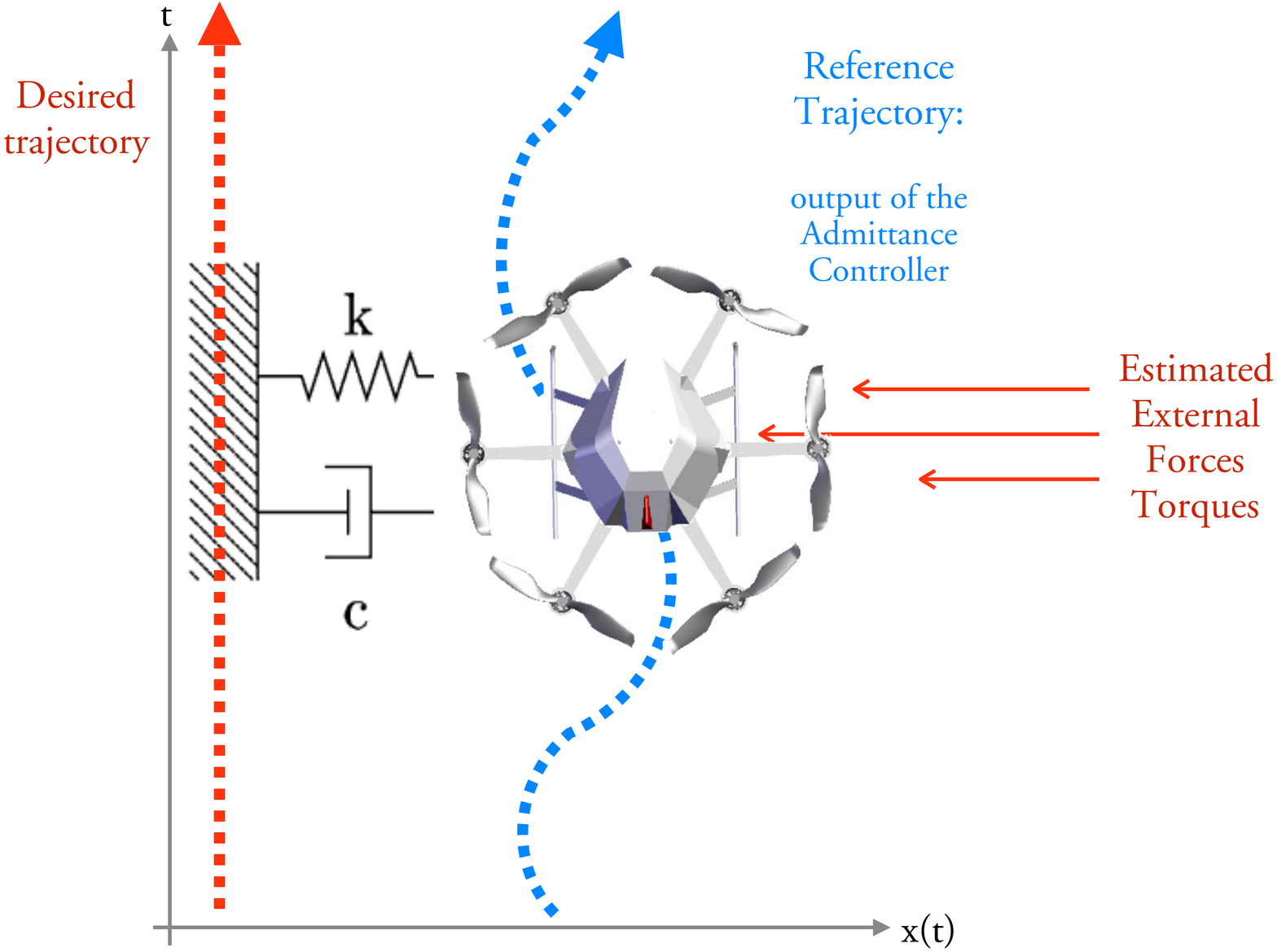}
	\caption{Schematic explanation of admittance control. Given a desired trajectory, the admittance controller generates a new reference trajectory according to the estimated external forces by simulating a second order \textit{spring-mass-damper} dynamic system.}
\label{pic:admc:intuitive}
\end{figure}

Admittance control is based on the idea that the trajectory which guarantees compliance with external force/torque is generated by simulating a \textit{spring-mass-damper} dynamic system. 
As represented in Figure \ref{pic:admc:intuitive}, the new trajectory is generated as if the \ac{MAV}, with virtual mass $m_v$, was connected to the desired trajectory via a spring with elastic constant $K$ and via a damper with damping coefficient $c$. The deviation from the desired trajectory is caused by the estimated external force/torque, which excites the virtual mass. 

Compliance with external force/torque can be tuned by changing $K$. Increasing values of $K$ simulate a stiffer spring, which in turn yields to better desired trajectory tracking. Conversely, $K = 0$ guarantees full compliance with the estimated external force.
	
\subsection{Trajectory Generation Law}
	The Admittance Controller generates a new reference trajectory independently for every axis. The trajectory generation law has been derived by the discretization of the differential equation \ref{eq:admc:xyzaxis} for the generation of the trajectory on $x_I$, $y_I$, $z_I$ and \ref{eq:admc:yaw} for the reference value of $\psi$.
    \small
	\begin{equation} \label{eq:admc:xyzaxis}
	m_v(\ddot{\Lambda}_{d, i} - \ddot{\Lambda}_{r,i})+c(\dot{\Lambda}_{d, i} - \dot{\Lambda}_{r,i}) + K(\Lambda_{d, i} -\Lambda_{r, i}) = F_{i}^{ext}
	\end{equation}
    \normalsize
	\begin{equation} \label{eq:admc:yaw}
	J_v(\ddot{\Lambda}_d - \ddot{\Lambda}_r)+c(\dot{\Lambda}_d - \dot{\Lambda}_r) +K(\Lambda_d -\Lambda_r) = \tau_z^{ext}
	\end{equation}

	The index $i$ denotes one of the axes $x_I$, $y_I$, $z_I$ , while  $\Lambda_{d,i}$ represents the desired  trajectory on the selected axis and $\Lambda_{r,i}$ the generated reference trajectory on the selected axis. The parameters $m_v$, $c$ and $K$ can be set independently for every axis and represent, respectively, the virtual spring, the virtual damping and the virtual mass of the controller. $J_v$ represents virtual inertia around the body $z$ axis. In addition, the controller only takes into account the desired pose by setting $\ddot{\boldsymbol{\Lambda}}_{d} = \dot{\boldsymbol{\Lambda}}_{d} = 0$, which is equivalent to assuming that the reference is only given in discrete steps.  
Additionally, in the output trajectory, we set  $\ddot{\boldsymbol{\Lambda}}_{r} = \dot{\boldsymbol{\Lambda}}_{r} = 0$ so that the controller only generates a reference for the pose of the \ac{MAV}. 	
\subsection{Robustness to Noise and Offsets}
If no force acts on the vehicle and the desired trajectory is constant, noise and other factors not included in the model (such as propeller efficiency) may result in a non-zero force estimate, which in turn causes a drift in the reference pose $\boldsymbol{\Lambda}_r$.Similarly, external factors such as constant wind or model-mismatches in the force and torque estimator may cause a constant offset in the estimated external disturbances.
\par
In order to avoid undesired drifts in the reference trajectory, the admittance controller has been integrated into a finite state machine (FSM), which monitors the magnitude of the force on each axis and decides whether to reject or take into account the estimated external force to compute a new trajectory reference \cite{Augugliaro2013}. Model mismatches and offset in the force/torque estimation can be taken into account by averaging the estimated external forces for a given time $T_{avg}$ while the helicopter is hovering. The computed offset is subtracted from each force/torque estimate produced by the force/torque estimator.
	
%As a safety feature, the controller saturates the estimate of external force/torque according to a predefined maximum value.
%In addition, a safety-box feature resets the reference trajectory to the current MAV estimated pose every time the controller tries to output a reference further than 1.5 m from the current MAV position. 
	
\subsection{Landing Detection and Auto-Disengagement}
	The controller must be engaged only once the vehicle is hovering, so that it does not take into account the ground reaction force as an external force resulting in an undesired reference trajectory as the output.
	The controller is also able to auto-disengage by detecting landing. Landing detection is achieved by checking the magnitude of the estimated external force and its angle $\theta_L$ with respect to the normal of the plane $x_I$,$y_I$; landing is detected when the magnitude of $\boldsymbol{F}^{ext}$ is bigger than a tunable threshold  $\bar{\boldsymbol{F}}^{ext}$  and if $\theta_L$ is smaller than a tunable threshold $\bar{\theta}_L$ for a given time $T_{Landing}$.

%%%%%%%%%%%%%%%%%%%%%%%%%%%%%%%%%%%%%%%%%%%%%%%%%%%%%%%%%%%%%%%%%%%%%%%%%%%%%%%%
% Evaluation
%%%%%%%%%%%%%%%%%%%%%%%%%%%%%%%%%%%%%%%%%%%%%%%%%%%%%%%%%%%%%%%%%%%%%%%%%%%%%%%%

\section{EVALUATION}
\label{sec:exp}
We now present four main experimental results. In the first two, we benchmark the performances of the force/torque estimator by comparing its estimates with the measurements of a force sensor and by showing its reaction speed in detecting a collision with a wall, respectively. In the third experiment, we test the admittance controller by interacting with the vehicle through a rope and finally, in the fourth experiment, we show the result of a payload carrying maneuver using two hexacopters.
\subsection{Force and torque estimator accuracy}
\begin{figure}
  \centering
  	\includegraphics[width=0.47\textwidth]{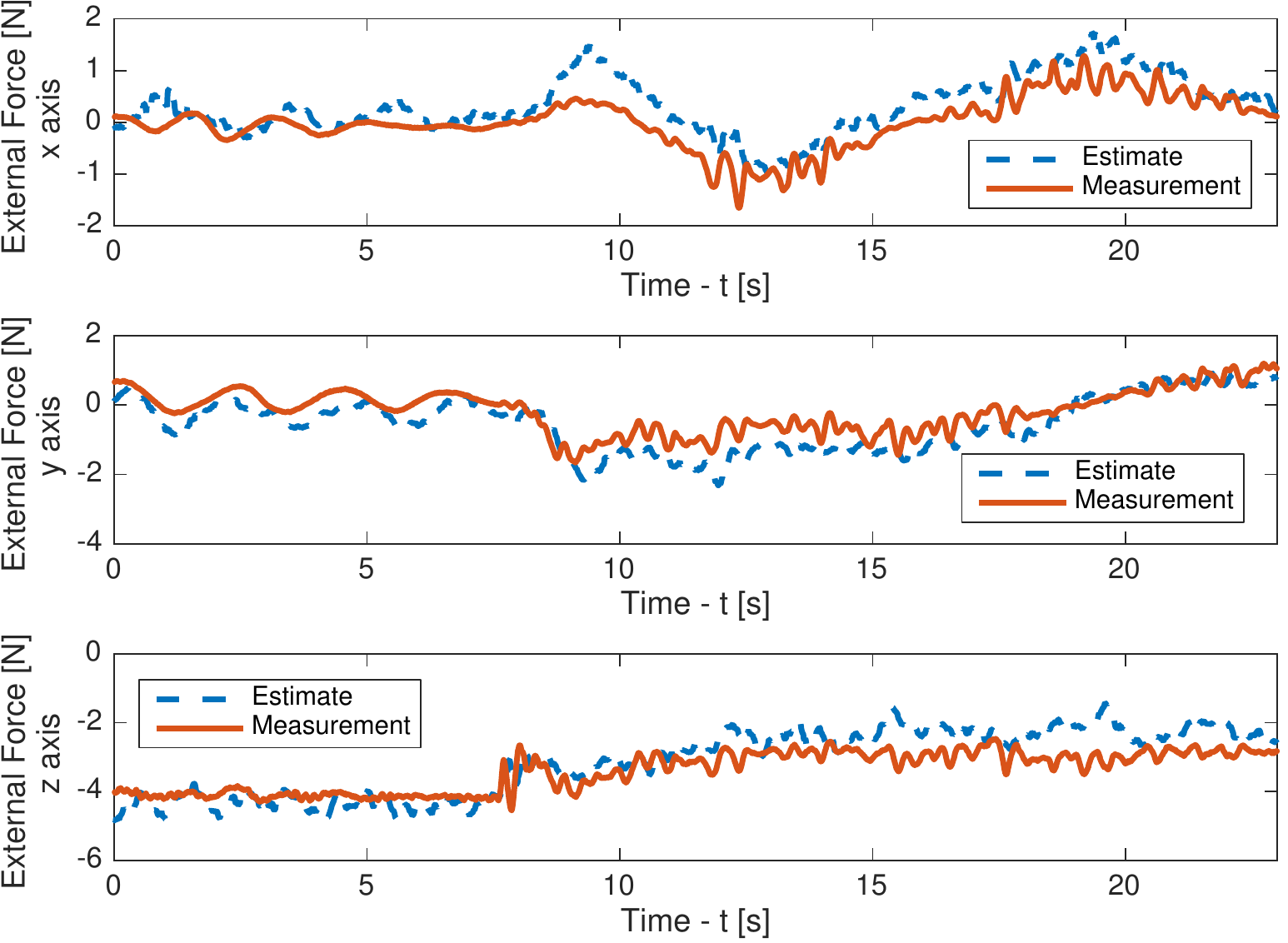}
  	\caption{Measured forces and estimated forces on $x_I$, $y_I$ and $z_I$ axis. We performed the experiment with the hexacopter on hover and using the visual inertial navigation system as state estimator.}
  	\label{pic:exp:ukf}
  \end{figure}

The first experiment was performed by connecting a Optoforce OMD-20-FE-200N force sensor at the bottom of the structure of the AscTech Firefly. An in-extensible nylon wire was attached to the sensor and interaction was performed by pulling the wire, while the \ac{MAV} was maintained steady at about $1.6$ m from the ground by the VI-odometry state estimator and MPC pose and attitude controller. The external forces due to the interaction measured and estimated on the $x_I$ and $y_I$ axis are represented in Figure \ref{pic:exp:ukf}. The RMS error between the measurements and the estimate is reported in Table \ref{tab:ukf:exp:1:error}.
     \begin{table}
   	\begin{center}
   		\caption{RMS estimation error for force/torque estimator validation} \vspace{1ex}
   		\label{tab:ukf:exp:1:error} 
   		\begin{tabular}{ccc}	
   			\hline
   			Disturbance & RMS error & Unit\\
   			\hline
   			${F}_x^{ext}$	& $0.4298$& N \\
   			${F}_y^{ext}$	& $0.4941$& N \\
   			${F}_z^{ext}$	& $0.5729$& N \\
   		\end{tabular}
   	\end{center}
   \end{table}
\subsection{Detecting Collision with a Wall}
In this experiment we use the force estimator to detect a collision with a wall and to send a "safe" new reference to the position and attitude controller. We use this experiment to highlight the reaction speed of the force estimator, showing that it is able to react in less than 20 ms. \par The logic for the  trajectory modification is  simply based on checking if the estimated external force exceeds a given threshold. If yes, than a safe reference is generated, which is computed by translating the impact point by a vector proportional to the impact force magnitude - computed at the step after the impact - and direction.
\begin{figure}
\centering
\includegraphics[width=0.37\textwidth]{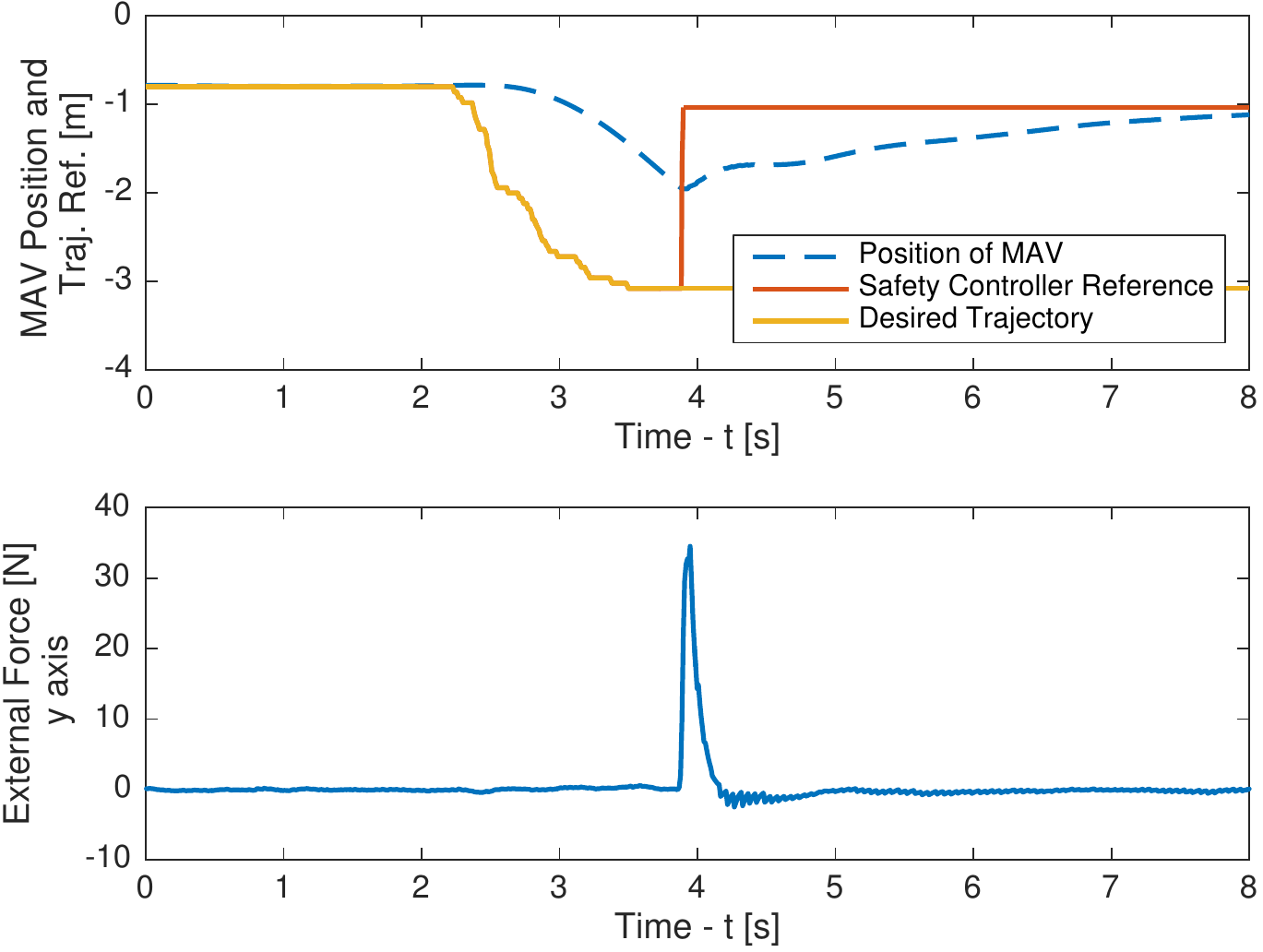}
    \caption{Wall collision detection experimental results. The wall is placed at $y_I=-1.95$ m. The impact generates an external force of $34.5$ N and a safe trajectory reference is generated 0.84 m further than the impact point. The impact happens at about $1.2$ m/s.}
\label{fig:exp:wall_collision}
\end{figure} 

\begin{figure}
  \centering
   \includegraphics[width=0.4
   \textwidth]{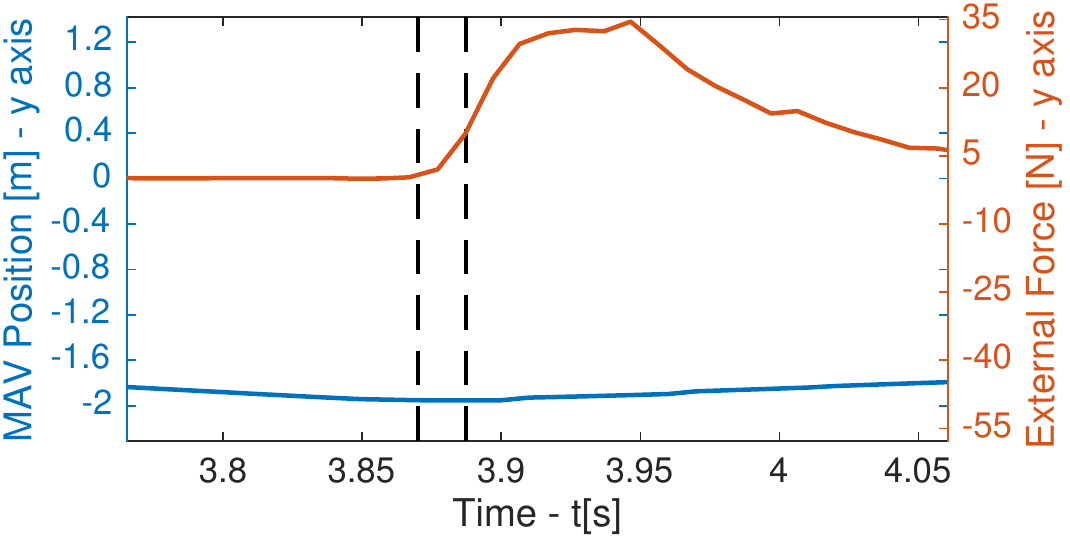}
 	\caption{Detail of the impact detection time computation. The vertical line on the left represents the impact instant $t=3.870$ s , while the vertical line on the right the detection instant $t=3.887$ s.}
\label{fig:exp:wall_collision_zoom}
\end{figure} 
\par In Figure \ref{fig:exp:wall_collision} we represent how the filter and the "safety" controller react in case of impact with a wall. Since the wall corresponds to the plane $y_I= -1.95$ m expressed in \textit{I} reference frame, we only show the $y$ component of
\begin{inparaenum}[(a)]
\item desired trajectory,
\item trajectory generated by safety controller,
\item pose of the \ac{MAV}, estimated using a motion capture system \cite{vicon}, and
\item estimated external force.
\end{inparaenum}

\subsection{Drone on a Leash}
\begin{figure}

\centering
\includegraphics[width=0.45\textwidth]{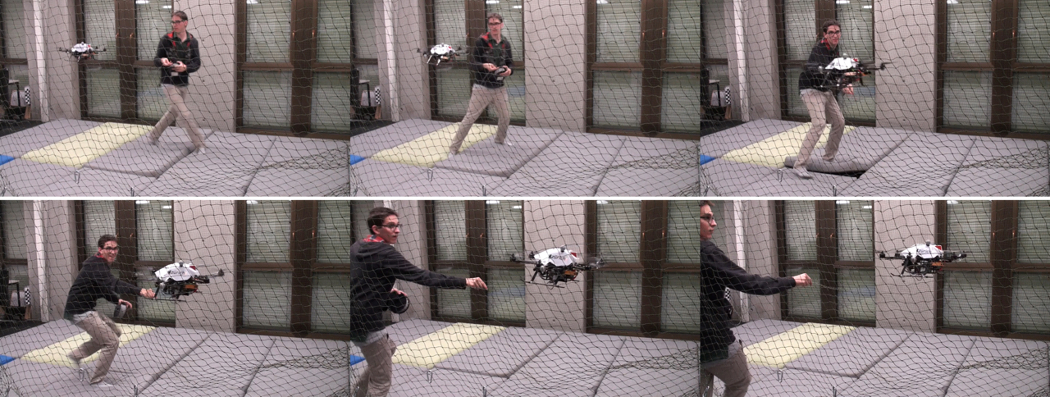}
\caption{Illustration of the human-hexacopter interaction. The interaction is performed through a string attached at the bottom of the \ac{MAV}. It is possible to experiment with different values of virtual spring, mass and damping to understand how they affect the behavior of the slave agent in a collaborative transportation task.}
\label{fig:admc:interaction}
\end{figure}
To test the capability of admittance controller we present a real-time interaction. A human operator can  interact with the MAV via a string attached at the bottom of the hexacopter. The results are represented in Figure \ref{fig:admc:interaction}.
  
\subsection{Cooperative payload transportation} 
Finally we present the cooperative transportation of a bulky object. The payload is a $1.2$ m carton tube weighing $0,37$ kg. The helicopters are two identical AscTec Firefly systems and the state estimator used is the visual-intertial navigation system (VI sensor). The master helicopter is controlled by a human operator, while the trajectory for the slave vehicle is generated by the admittance controller. Both the helicopters are attached at the extremities of the payload via a nylon wire, which cancels any torque. The admittance controller is used to generate a compliant trajectory on the $x_I$ and $y_I$ axis, while tracking a given altitude reference on $z_I$. We achieved this by setting the virtual spring value on $x$ and $y$ axis to be $K=0$ N/m, while on the $z$ a stiff virtual spring of value $K=10$ N/m has been used.
\par
  
  \begin{figure}
  	\centering
  	\includegraphics[width=0.47\textwidth]{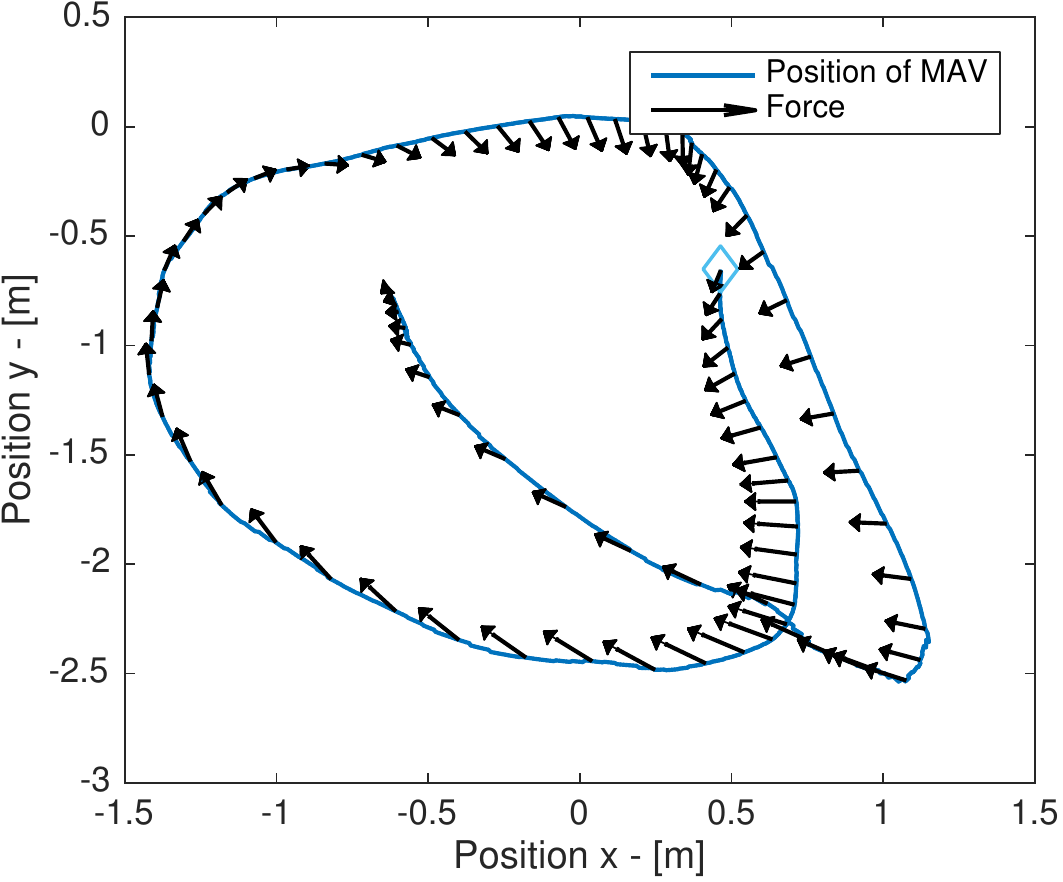}
  	\caption{Executed trajectory by the slave agent. The estimated external forces acting on the slave agent is depicted by the black arrows.}
  \label{pic:exp:2dronestraj}
  \end{figure}
\begin{figure}
\centering
\includegraphics[width=0.47\textwidth]{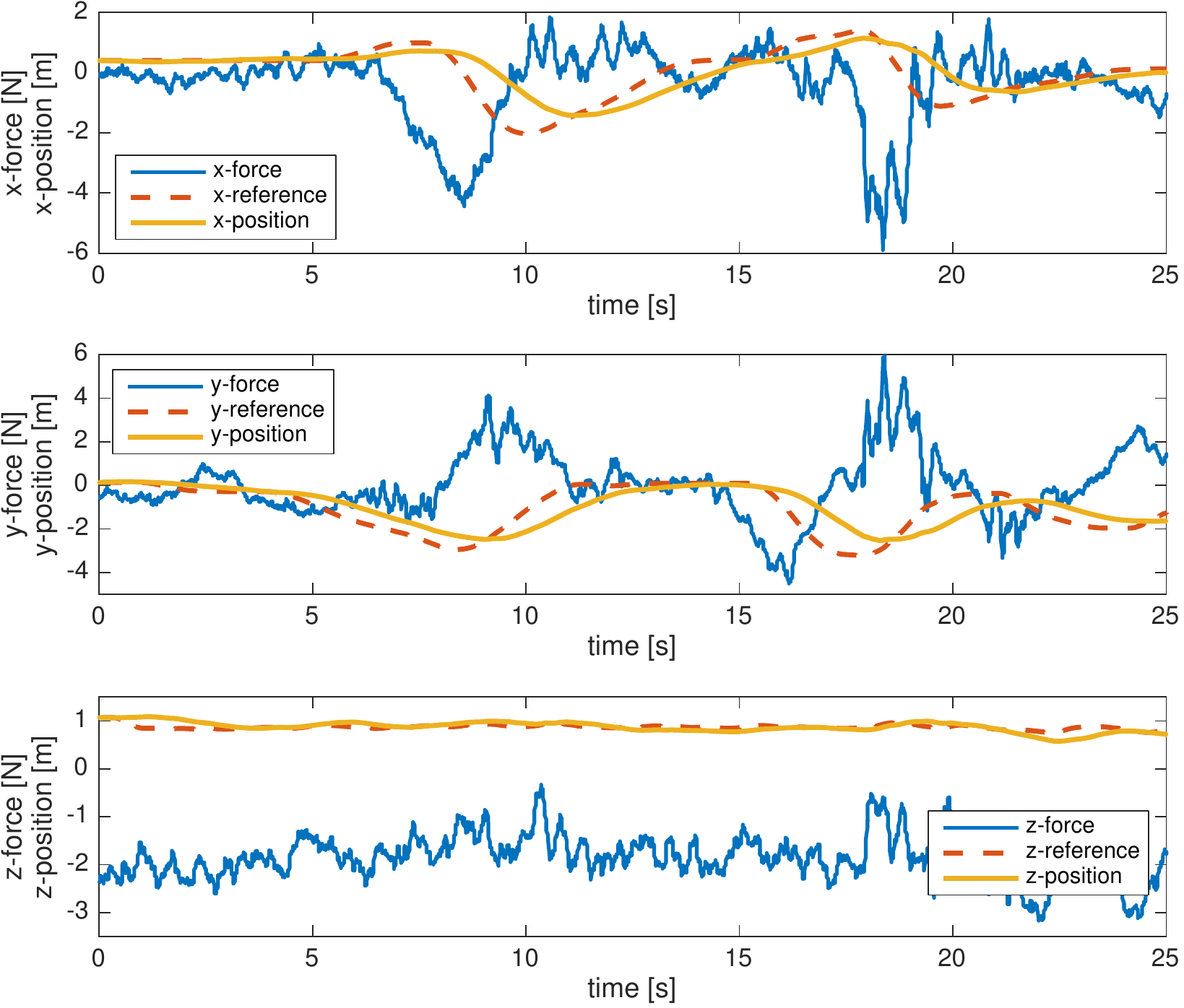}
\caption{Estimated external force, admittance controller reference trajectory and estimated position of the slave agent during the collaborative transportation.}
\label{pic:exp_2dronesforcetraj}
\end{figure}
\begin{figure}
\centering
\includegraphics[width=0.47\textwidth]{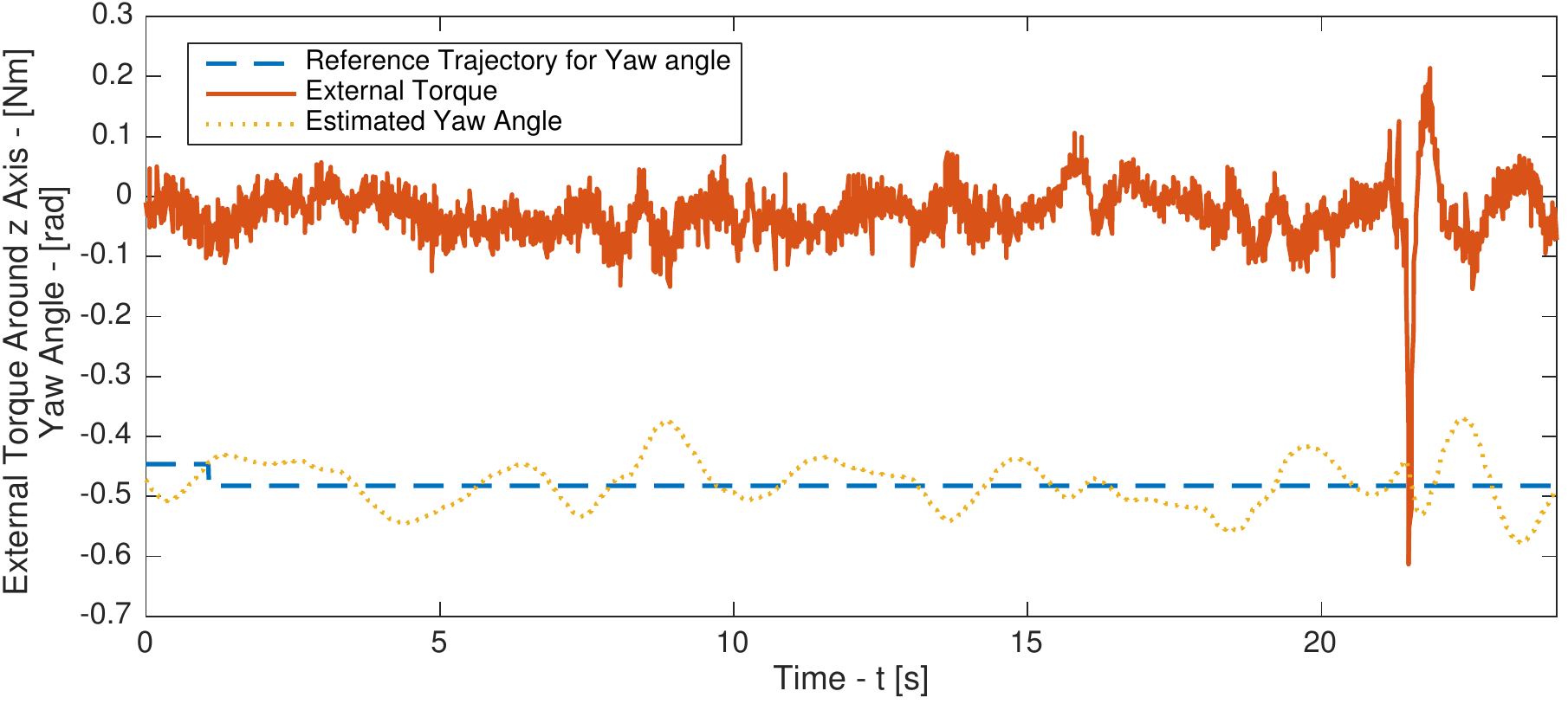}
\caption{Estimated external torque around $z$ axis, reference trajectory for yaw attitude angle, and estimated yaw attitude angle (VI-sensor). The peak in torque estimation at $t = 22$ s corresponds to the accidental collision of one of the propellers of the \ac{MAV} with the payload.}
\label{pic:exp:torqueexp}
\end{figure}

The setup of the experiment is represented in Figure \ref{pic:exp:masterslave}. The reference trajectory for the slave and its position are represented in Figure \ref{pic:exp:2dronestraj}, while the reference trajectory and the estimated external forces for every axis are shown in Figure \ref{pic:exp_2dronesforcetraj}. Finally, the estimated external torque as well as the yaw angle are displayed in Figure \ref{pic:exp:torqueexp}.
    
%%%%%%%%%%%%%%%%%%%%%%%%%%%%%%%%%%%%%%%%%%%%%%%%%%%%%%%%%%%%%%%%%%%%%%%%%%%%%%%%
% Conclusion
%%%%%%%%%%%%%%%%%%%%%%%%%%%%%%%%%%%%%%%%%%%%%%%%%%%%%%%%%%%%%%%%%%%%%%%%%%%%%%%%

\section{CONCLUSION}
\label{sec:futwrk}
In this work we showed that collaborative transportation of bulky payload can be achieved without relying on a communication network between the involved agents. We achieved this by making use of an admittance controller in conjunction with a force estimator based on the Unscented Kalman Filter. The force estimates are obtained using the state information provided by a visual inertial navigation system. We include demonstration of the filter performance by detecting wall collision and real time interaction with a human.

%%%%%%%%%%%%%%%%%%%%%%%%%%%%%%%%%%%%%%%%%%%%%%%%%%%%%%%%%%%%%%%%%%%%%%%%%%%%%%%%
%%%%%%%%%%%%%%%%%%%%%%%%%%%%%%%%%%%%%%%%%%%%%%%%%%%%%%%%%%%%%%%%%%%%%%%%%%%%%%%%

\addtolength{\textheight}{-12cm}   % This command serves to balance the column lengths

%%%%%%%%%%%%%%%%%%%%%%%%%%%%%%%%%%%%%%%%%%%%%%%%%%%%%%%%%%%%%%%%%%%%%%%%%%%%%%%%

%%%%%%%%%%%%%%%%%%%%%%%%%%%%%%%%%%%%%%%%%%%%%%%%%%%%%%%%%%%%%%%%%%%%%%%%%%%%%%%%

%%%%%%%%%%%%%%%%%%%%%%%%%%%%%%%%%%%%%%%%%%%%%%%%%%%%%%%%%%%%%%%%%%%%%%%%%%%%%%%%
\bibliographystyle{IEEEtran}

\bibliography{bibliography/Mendeley,bibliography/websites}

\begin{acronym}
\acro{TRADR}{``Long-Term Human-Robot Teaming for Robots Assisted Disaster Response''}
\acro{MAV}{Micro Aerial Vehicle}
\acro{MBZIRC}{Mohamed Bin Zayed International Robotics Challenge}
\acro{UAV}{Unmanned Aerial Vehicle}
\acro{UKF}{Unscented Kalman Filter}

\end{acronym}

\end{document}